\documentclass[10pt, a4paper]{article}
\usepackage{times}
\usepackage{latexsym}
\usepackage{CJKutf8}
\usepackage[T1]{fontenc}
\usepackage[utf8]{inputenc}
\usepackage{microtype}
\usepackage{inconsolata}
\usepackage{booktabs}
\usepackage{graphicx} 
\usepackage{hyperref}
\usepackage{inconsolata}
\renewcommand\footnotemark{}

\usepackage{tabularx} 
\usepackage{booktabs} 
\usepackage{makecell} 
\usepackage{multirow} 

\usepackage[final]{lrec-coling2024} 

\title{MCTS: A Multi-Reference Chinese Text Simplification Dataset}

\name{
        Ruining Chong\textsuperscript{1}, 
        Luming Lu\textsuperscript{1}, 
        Liner Yang\textsuperscript{1\dag} \thanks{\textsuperscript{\dag}Corresponding author: Liner Yang.}, 
        Jinran Nie\textsuperscript{1*}, 
        \thanks{\textsuperscript{*}The work was completed during their studies at Beijing Language and Culture University.}\\
\large{\textbf{
        Zhenghao Liu\textsuperscript{2}, 
        Shuo Wang\textsuperscript{3}, 
        Shuhan Zhou\textsuperscript{1*}, 
        Yaoxin Li\textsuperscript{1*} and 
        Erhong Yang\textsuperscript{1}}}} 

\address{
\textsuperscript{1}School of Information Science, Beijing Language and Culture University \\
\textsuperscript{2}Department of Computer Science and Technology, Northeastern University \\ 
\textsuperscript{3}Department of Computer Science and Technology, Institute for AI, Tsinghua University \\
 lineryang@gmail.com\\
}

\abstract{
Text simplification aims to make the text easier to understand by applying rewriting transformations.
There has been very little research on Chinese text simplification for a long time.
The lack of generic evaluation data is an essential reason for this phenomenon.
In this paper, we introduce MCTS, a multi-reference Chinese text simplification dataset. 
We describe the annotation process of the dataset and provide a detailed analysis.
Furthermore, we evaluate the performance of several unsupervised methods and advanced large language models.
We additionally provide Chinese text simplification parallel data that can be used for training, acquired by utilizing machine translation and English text simplification.
We hope to build a basic understanding of Chinese text simplification through the foundational work and provide references for future research. All of the code and data are released at \url{https://github.com/blcuicall/mcts/}.
 \\ \newline  }

\begin{document}
\begin{CJK*}{UTF8}{gkai}

\maketitleabstract
\section{Introduction}

The task of text simplification aims to make the text easier to understand by performing multiple rewriting transformations. 
It can provide reading assistance for children \cite{Kajiwara:13}, non-native speakers \cite{Paetzold:16}, and people with language disorders \cite{Carroll:98,Paetzold:16,Evans:14e}. 
Moreover, text simplification can also be used as a data augmentation method to benefit downstream natural language processing (NLP) tasks \cite{Van:21}.

For a long time, the research of text simplification systems mainly depends on large-scale parallel corpora for training, such as WikiLarge \citeplanguageresource{zhang:17} and Newsela \citeplanguageresource{Xu:15}. 
But due to the limitation of existing data in language and domain, recent work on text simplification systems has started to focus on unsupervised methods and achieves good results \cite{surya:19,kumar:20,martin:22}, which makes it possible to build Chinese text simplification systems independent of large-scale parallel corpora. 
In this case, evaluating the Chinese text simplification systems becomes a problem that needs to be solved.
On the other hand, large language models have the ability to solve various NLP tasks \cite{thoppilan:22,chowdhery:22}. 
Recently a series of large language models represented by ChatGPT \footnote{\url{https://chat.openai.com/chat}} perform well on many tasks \cite{qin2023chatgpt,jiao2023chatgpt,bang2023multitask}.
In English text simplification, Feng et al. \shortcite{feng:23} find that large language models outperform state-of-the-art methods and are judged to be on par with human annotators.
Nevertheless, whether these models can achieve the same excellent results in Chinese text simplification remains unclear.

To solve these problems, we introduce MCTS, a multi-reference dataset for the evaluation of Chinese text simplification models. 
MCTS comprises 3,615 human simplifications associated with 723 original sentences selected from the Penn Chinese Treebank \citeplanguageresource{xue:05} (5 simplifications per original sentence), and it encompasses a diverse array of rewriting transformations.
To our knowledge, it is the first published multi-reference Chinese text simplification evaluation dataset.
We aim for this dataset to serve as a gauge for assessing the current state of Chinese text simplification and to furnish benchmarks that will inform and guide future scholarly inquiries.

We have also designed several simple, unsupervised Chinese text simplification methods and evaluated them using our newly proposed dataset. 
These approaches are intended to establish a foundation for subsequent research endeavors.
Moreover, we assess the capabilities of advanced large language models, such as ChatGPT, in performing Chinese text simplification. 
Our findings indicate that these large language models surpass our established unsupervised baselines.
Nonetheless, there remains a discernible disparity in quality compared to human-crafted simplifications.
In summary, our contributions are listed below:
\begin{itemize}
\item We manually annotate a dataset that can be used for the evaluation of Chinese text simplification.
It is a multi-reference dataset that encompasses multiple types of rewriting transformations.
To our knowledge, it is the first published multi-reference Chinese text simplification evaluation dataset.
\item We provide several text features and conduct a detailed analysis of the dataset, which could help to understand the characteristics of human Chinese text simplification.
\item On the proposed dataset, we evaluate the performance of several unsupervised methods and advanced large language models, which could serve
as the baselines for future research.
\item As a by-product, we additionally release the Chinese text simplification parallel dataset suitable for training purposes, obtained through machine translation and the English text simplification model. The dataset comprises 691,474 sentence pairs. To our knowledge, it is the largest Chinese text simplification training dataset.

\end{itemize}

\section{Related Work}

\subsection{Evaluation Data for English Text Simplification}
In the field of English text simplification, the research on system evaluation has accumulated rich experience and results, providing valuable references and insights for the construction of Chinese text simplification evaluation data.

Early evaluation data for English text simplification mainly consist of sentence pairs obtained from English Wikipedia and Simple English Wikipedia through automatic sentence alignment.
However, the Simple English Wikipedia was argued to contain many inadequate or inaccurate simplifications \cite{yasseri:12, Xu:15}. 
Furthermore, evaluation data with only a single reference is insufficient to ensure reliability and comprehensiveness in highly subjective text simplification tasks.

For the above reasons,  \citetlanguageresource{xu:16} introduced TurkCorpus, a multi-reference dataset for evaluating English text simplification.
They first collected 2,359 original sentences from English Wikipedia and then obtained 8 manual reference references for every original sentence via crowdsourcing.
The dataset can be used for evaluation metrics requiring multiple references, such as BLEU \cite{papineni:02} and SARI \cite{xu:16}.
However, the rewriting transformations involved in TurkCorpus are simplistic. Annotators were asked to simplify a sentence mainly by lexical paraphrasing but without deleting content or splitting the sentences. 
Another multi-reference dataset for English text simplification, HSplit \citeplanguageresource{sulem:18}, uses the same original sentences from the TurkCorpus test set but only focuses on sentence splitting.

The simplicity of the rewriting transformations hinders research into the model's ability to perform a wide range of rewriting operations.
To solve this problem, \citetlanguageresource{alva:20} created the ASSET dataset.
Using the same origin sentences, They extended TurkCorpus through crowdsourcing. 
They defined a wider range of rewriting transformations, with lexical paraphrasing (lexical simplification and reordering), sentence splitting, and compression (deleting unimportant information).
Through experimentation, it was discovered that the ASSET dataset features more abstractive simplifications, which are deemed simpler compared to those found in other evaluation corpora.
Now ASSET has been adopted as a standard dataset for evaluating English text simplification systems.

Similar to ASSET, MCTS is a dataset with multiple references and multiple rewriting transformations. 
To our knowledge, this dataset is the largest and most comprehensive evaluation dataset for the Chinese text simplification task to date.

\begin{table*}[t!]
\small
\begin{center}
\begin{tabular}{cl}
\toprule
\bf Original & 为适应大西南进出口物资迅速增长的需要，北部湾沿海四市开始了新一轮建港热潮。 \\
            & In order to adapt to the rapid growth of import and export materials in the southwest, four\\& coastal cities in the Beibu Gulf have started a new wave of port construction.\\
\bf Reference & 大西南进口和出口物资的需要迅速增长，北部湾沿海四座城市兴起了新一轮港口建设。  \\
            & The demand for imported and exported materials in the southwest has grown rapidly. Four\\& coastal cities in the Beibu Gulf have begun a new round of port construction. \\
\midrule 
\bf Original & 中国又一条煤炭运输大通道──连接天津蓟县与天津港之间的蓟港铁路日前破土动工。 \\
            & Another major coal transportation corridor in China - the Jigang Railway connecting Tianjin\\& Jixian County and Tianjin Port - has recently broken ground. \\
\bf Reference & 蓟港铁路连接天津蓟县和天津港，用于煤炭运输，前几天开始建造。  \\
            & Jigang Railway connects Tianjin Jixian County and Tianjin Port for coal transportation, and \\& construction began a few days ago. \\
\midrule 
\bf Original & 按设计，“进步M-24”号载重飞船可同轨道站在无人操纵的情况下进行自动对接。 \\
            & According to the design, the "Progress M-24" heavy-duty spacecraft can automatically dock \\&  with the orbital station under unmanned control. \\ 
\bf Reference & 按照设计，无人操控的“进步M-24”号飞船可以自动对接轨道站。  \\
            & According to the design, the unmanned "Progress M-24" spacecraft can automatically dock \\& with the orbital station. \\
\bottomrule
\end{tabular}
\end{center}
\caption{\label{table1} Examples of simplifications collected for MCTS }
\end{table*}

\subsection{Unsupervised Text Simplification}
Unsupervised text simplification methods do not require aligned complex-simple sentence pairs.
After Surya et al. \shortcite{surya:19} introduced the first unsupervised neural text simplification system by importing adversarial and denoising auxiliary losses, the field of unsupervised text simplification has rapidly developed.

Currently, the research on unsupervised text simplification mainly adopts two strategies.
One approach is to train models using text mining or automatically generated pseudo-data pairs, and the other is to adopt cross-language transfer learning methods for text simplification. 

MUSS \cite{martin:22} is an outstanding representative of the first methods class.
It could effectively generate simplified text by using text paraphrasing data obtained through data mining and realize the difficulty control with ACCESS \cite{martin:20controllable} during inference.
Brunato et al.\shortcite{brunato-2016-paccss} mined pseudo-parallel simplification data from a wide amount of texts extracted from the web using metrics such as semantic similarity and readability.
Lu et al.\shortcite{lu2021unsupervised} constructed pseudo-parallel data for text simplification from large-scale machine translation data pairs.

The second category of methods, cross-language transfer learning, explores how to transfer the knowledge of English text simplification to other languages.
Mallinson et al.\shortcite{mallinson-2020-zero} used a unique model architecture and multi-task learning strategy to realize cross-language zero-shot learning for German text simplification.

In this study, we construct several unsupervised models for Chinese text simplification using pseudo-data construction and cross-language transfer learning as the main ideas, which can provide important baseline references for future research work.



\section{Creating MCTS}
In this section, we describe MCTS in more detail.
In section \ref{3.1}, we introduce the preparation of original sentences.
In section \ref{3.2}, we introduce the annotation process of MCTS.

\subsection{Data Preparation}\label{3.1}
We choose Penn Chinese Treebank (CTB) as the source of the original sentence in the dataset. 
CTB is a phrase structure tree bank built by the University of Pennsylvania. 
It includes Xinhua news agency reports, government documents, news magazines, broadcasts, interviews, online news, and logs.
We concentrate on news content because of its wide accessibility and significant influence.
The rich corpus in the news area also makes it conducive to further work advancement.

We initially employ a filter based on the HSK\footnote{\url{https://www.chinesetest.cn/}} average vocabulary difficulty, 
setting a threshold for the average vocabulary level to filter out simple sentences, 
ensuring the selected original sentences possess sufficient complexity. 
Subsequently, through manual screening of the remaining sentences, we selected 723 sentences to serve as the original sentences for our study.

\subsection{Annotation Process} \label{3.2}
MCTS stands as a fully manually annotated evaluation dataset. Here is a detailed description of the annotation process.

\paragraph{Annotator Recruitment}
All the annotators we recruited are native Chinese speakers and are undergraduate or graduate students in school. 
Most of them have a background in linguistics or computer science.
All annotators need to attend a training course and take the corresponding Qualification Test (see more details below) designed for our task.
Only those who have passed the Qualification Test could enter the Annotation Round.

\paragraph{Simplification Instructions}
We provide the exact instructions for annotators for the Qualification Test and the Annotation Round. 
In the instructions, we define three types of rewriting transformations. 
\begin{itemize}
\item \textit{Paraphrasing}: Replacing complex words or phrases with simple formulations.
\item \textit{Compression}: Deleting repetitive or unimportant information from the sentence.
\item \textit{Structure Changing}: Modifying complex sentence structures into simple forms.

\end{itemize}

Compared to rewriting transformations involved in ASSET, we have made the following two changes：
(1) We employ \textit{Paraphrasing}, which extends beyond the narrow scope of mere \textit{Lexical Paraphrasing}.
The paraphrasing transformation in Chinese is much more flexible than in English.
It not only includes synonym substitution but also covers the interpretation of complex phrases or idioms.
(2) We replaced \textit{Sentence Splitting} with \textit{Structural Changing}. 
In Chinese text simplification, it's more common to convert complex sentences into compound sentences with several clauses separated by commas rather than splitting them into separate sentences with periods. 
The narrowly defined Sentence Splitting is not well-suited to this situation.
The category of \textit{Structure Changing} encapsulates a variety of shifts, including but not limited to the conversion from passive to active voice, the simplification of double negatives into affirmative constructions, the substitution of synonyms, the interpretation of complex phrases or idioms, and the rephrasing of rhetorical questions into declarative statements.

These two changes ensure that the defined rewriting transformations are both non-redundant and exhaustive, and they align with the practical context of Chinese text simplification.

For every rewriting transformation, we provide several examples. 
Annotators could decide for themselves which types of rewriting to execute in any given original sentence.

\paragraph{Qualification Test}
At this stage, we provided 20 sentences to be simplified.
Annotators needed to simplify these sentences according to the instructions given. 
We checked all submissions to filter out annotators who could not perform the task correctly.
Of the 73 people who initially registered, only 35 passed the Qualification Test (48\%) and worked on the task.

\paragraph{Annotation Round}
Annotators who passed the Qualification Test had access to this round.
To facilitate annotating work, we provided a platform that can display the difficulty level of words in a text.
We collected 5 simplifications for each of the 723 original sentences. 
Table \ref{table1} presents a few examples of simplifications in MCTS, together with their English translation.

\begin{figure*}
    \centering
    \includegraphics[width=\linewidth]{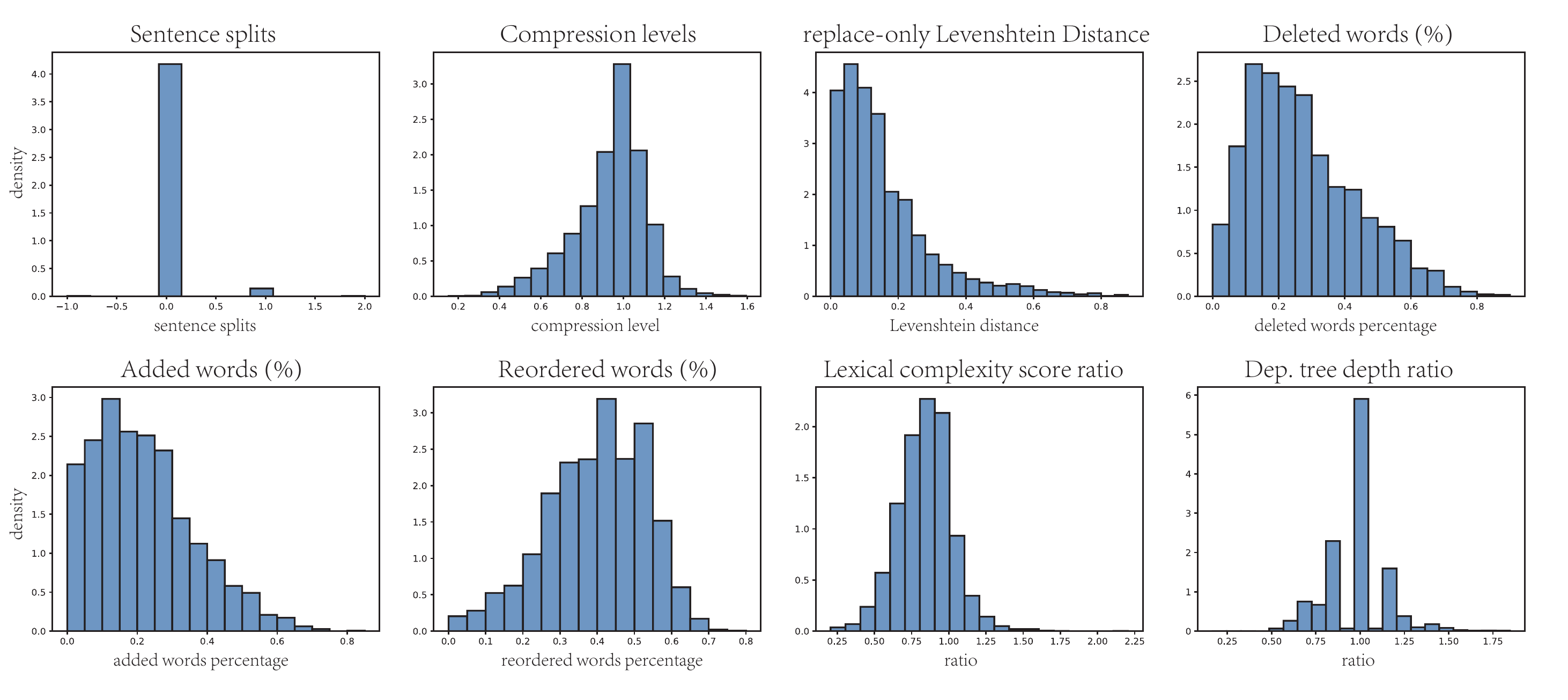}
    \caption{Density of text features in simplifications from MCTS}
    \label{analyze}
    \vspace*{-0.3cm}
\end{figure*}

\section{Dataset Analysis}
Following ASSET \cite{alva:20}, we analyze various text features within the MCTS dataset to examine the nature of its simplifications comprehensively.
We provide a detailed introduction in sections \ref{4.1} and \ref{4.2}. 
In section \ref{4.3}, we report a detailed comparison between MCTS and another recently proposed dataset.

\subsection{Text Features} \label{4.1}
We calculate several low-level features for all simplification examples to measure the rewriting transformations included in MCTS.
These features are listed below.

\begin{itemize}
    \item Number of sentence splits: The difference between the number of sentences in the simplification and the number of sentences in the original sentence.
    
    \item Compression level: The number of characters in the simplification divided by the number of characters in the original sentence.
    
    \item Replace-only Levenshtein distance: The character-level Levenshtein distance \cite{levenshtein1966binary} for replace operations only divided by the length of the shorter string in the original sentence and simplification. As described in ASSET, ignoring insertions and deletions can make this feature independent of compression level and serve as a proxy for measuring the lexical paraphrases of the simplification.
    
    \item Proportion of words deleted, words added and words reordered: The number of words deleted/reordered from the original sentence divided by the number of words in the original sentence; and the number of words that were added to the original sentence divided by the number of words in the simplification.

    \item Lexical complexity score ratio: We compute the score as the mean squared lexical difficulty level in HSK. The ratio is then the value of this score on the simplification divided by that of the original sentence, which can be considered as an indicator of lexical simplification.

    \item Dependency tree depth ratio: The ratio of the depth of the dependency parse tree of the simplification relative to that of the original sentence. Following ASSET \cite{alva:20}, we perform parsing using spaCy \footnote{\url{https://github.com/explosion/spaCy}}. This feature can reflect structural simplicity to a certain extent.
    
\end{itemize}

\subsection{Results and Analysis}  \label{4.2}
The density of all these features is shown in Figure \ref{analyze}.
We can see that sentence splitting operation appears not frequently on MCTS.
By observing the data, we believe this is due to the characteristics of the Chinese language.
Compound sentences are commonly used in Chinese, and one sentence consists of two or more independent clauses.
As we mentioned in section \ref{3.2}, during the simplification, annotators tend to rewrite a complex sentence with nested clauses into compound sentences rather than multiple simple sentences.
So this is not to say that Chinese text simplification rarely involves sentence structure change, but that the way of structural change is not limited to sentence splitting.

Although we have introduced compression as a rewriting transformation in the simplification instructions, the compression ratio is not too concentrated on the side less than 1.0.
The reason is that, on the one hand, the annotators tend to retain as much semantic information as possible, and on the other hand, more characters may be added when paraphrasing.

By analyzing replace-only Levenshtein distance, we can see that the simplifications in MCTS have a considerable degree of paraphrasing the input as simplifications are distributed at all levels.
Regarding the distribution of deleted, added, and reordered words, we can find that the peaks all occur at positions greater than 0.0. 
This further reveals the plentiful rewriting operations contained in MCTS.

In terms of lexical complexity, we can clearly see the high density of ratios less than 1.0, indicating that simplification has significantly lower lexical complexity compared to the original sentence.
Some instances have a lexical complexity ratio greater than 1.0, which may be due to deleted simple words in the process of sentence compression.

Finally, the dataset shows a high density of a 1.0 ratio in dependency tree depth. This may indicate that significant structural changes were not made.

\subsection{Comparison with CSS}  \label{4.3}

We have also noticed another recent work by \citet{yang2023new}, which also attempts to construct a multi-reference Chinese text simplification evaluation dataset independently. 
However, their work is not sufficiently comprehensive.
The created CSS dataset lacks a validation set, and each original sentence has only two references, constraining its universality.
We report a detailed comparison of two datasets in Tables \ref{table:comparison} and \ref{table:hskcomp}.
We use the sentence-transformers toolkit \cite{reimers2020making} to calculate semantic similarity and use the python-Levenshtein package to calculate edit distance.

In table \ref{table:comparison}, we observe that MCTS significantly surpasses CSS in terms of data volume, both in the number of references (5 vs. 2) and the quantity of source sentences (723 vs. 383).
By analyzing the sentence length, we can find that MCTS is slightly longer than CSS in terms of the original sentence length.
From the length changes before and after simplification, it can be seen that MCTS is more inclined to perform vocabulary level replacement or deletion operations compared to CSS.
Finally, by analyzing semantic similarity and editing distance, we find that MCTS has made greater simplification operations than CSS.

In table \ref{table:hskcomp}, we measure the difference between two datasets in lexical complexity using the HSK Level \cite{kong2022multitasking} metric. 
We provide a detailed introduction to this metric in section \ref{5.2}
In simple terms, the higher the proportion of words in levels 1-3 (7+), the easier (more challenging) the text is understood.
We can find that the original text of MCTS uses more difficult vocabulary compared to CSS.
While observing the simplified version, MCTS is simpler than CSS.
This is enough to demonstrate that MCTS has undergone a more in-depth simplification.

\begin{table*}[t]
	\begin{center}
     \setlength{\tabcolsep}{8pt}
    \renewcommand\arraystretch{1.2}
	\begin{tabular}{c|c|c|c|c|c|c|c|c}
		\hline
		\multirow{2}{*}{Datasets} & \multirow{2}{*}{NRef.} & \multirow{2}{*}{Sent.} & \multicolumn{2}{c|}{Length (Char)} & \multicolumn{2}{c|}{Length (Token)} & \multirow{2}{*}{SS} & \multirow{2}{*}{LD} \\ \cline{4-7}
		& & & Ori.       & Ref.      & Ori.       & Ref.       &  &       \\ 
 \hline
		CSS               & 2                      & 383                 & 49.49               & 47.3                & 31.37               & 28.06              & 93.03             & 17.15        \\ 
  		MCTS              & 5                      & 723                 & 50.13               & 50.05               & 32.6                & 30.04              & 90.13             & 27.33        \\
		\hline
	\end{tabular}
\end{center}

\caption{Comparison of the MCTS and CSS datasets.
NRef. refers to the number of simplified sentences corresponding to each original sentence. 
Sent. refers to the number of original sentences. Length (Char) and Length (Token) refer to the average character length and the average token length of each sentence, respectively. 
SS refers to Semantic Similarity. 
LD refers to Levenshtein Distance.
Ori. represents the score of the original sentences, and Ref. represents the score of the reference sentences. }
\label{table:comparison}
\end{table*}

\begin{table*}[htbp]
	\centering
    \setlength{\tabcolsep}{10pt }
    \renewcommand\arraystretch{1.2}
	\begin{tabular}{c|c|c|c|c}
		\hline
		\multirow{2}{*}{Datasets} 
		& \multicolumn{2}{c|}{Ori.} & \multicolumn{2}{c}{Refs.} \\ 
		\cline{2-5}
		& L1-3(\%) & L7+(\%) & L1-3(\%) & L7+(\%) \\ \hline
  	CSS & 40.93     & 44.74      & 44.3      & 41.79     \\ 
		MCTS & 39.64     & 45.05      & 45.85    & 39.65    \\ 
		\hline
	\end{tabular}
	\caption{
                    Comparison of MCTS and CSS on HSK Level Scores. 
                    L1-3 and L7+, respectively, represent the proportion of vocabulary at levels 1-3 and levels 7+.
                }
	\label{table:hskcomp}
\end{table*}

\section{Experiment}

In order to measure the development status of Chinese text simplification and provide references for future research, we conduct a series of experiments on the proposed MCTS.

\subsection{Methods}
We attempt several unsupervised Chinese text simplification methods and large language models and provide their results on MCTS.
The first three are unsupervised methods that utilize automatic machine translation technology. 
We use Google Translator \footnote{\url{https://translate.google.com/}} to translate.
These unsupervised methods can be used as the baselines for future work.

We divide all the 723 sentences in MCTS into two subsets: 366 for validation and 357 for testing the Chinese text simplification models. 
For those model training methods, the valid set is used to select the best-performing checkpoints, and the final score is calculated on the test set.

\paragraph{Direct Back Translation}
As high-frequency words tend to be generated in the process of neural machine translation \cite{lu2021unsupervised}, back translation is a potential unsupervised text simplification method.
We translate the original Chinese sentences into English and then translate them back to obtain simplified results.
We choose English as the bridge language because of the rich bilingual translation resources between Chinese and English.

\paragraph{Translated Wiki-Large}
Translating existing text simplification data into Chinese is a simple way to construct pseudo data.
We translated English sentence pairs in Wiki-Large into Chinese sentence pairs and used them to train a Chinese version BART-based \cite{shao-2021-cpt} model as one of our baselines. 

We have followed the concept of previous work \cite{martin:22} and uniformly described similar methods for constructing pseudo data as unsupervised methods.

\begin{figure*}[t!]
    \centering
    \includegraphics{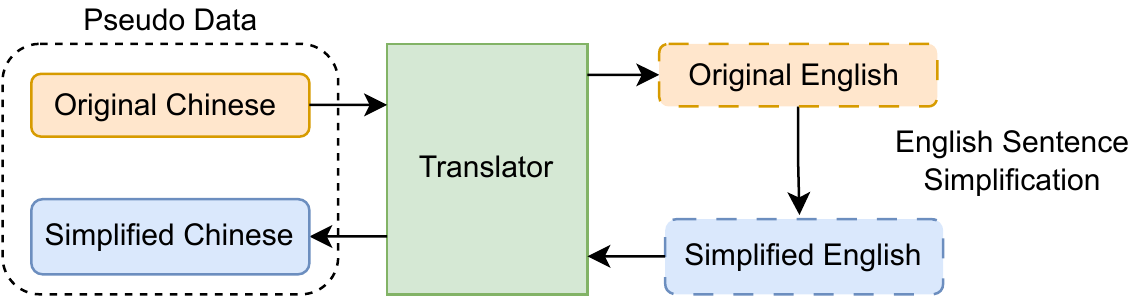}
    \caption{Pseudo data acquisition process}
    \label{pesdo}
    \vspace*{-0.3cm}
\end{figure*}

\paragraph{Cross-Lingual Pseudo Data}
In addition to the above two methods, we also design a way to construct pseudo data for Chinese text simplification, which can leverage the knowledge from English text simplification models.
As shown in Figure \ref{pesdo}, we first collect a large amount of Chinese sentence data, for example, the People's Daily Corpus. 
Then, we translate these sentences into English and simplify them using existing English text simplification models.
Finally, we translate the simplified English sentences back into Chinese and align them with the original Chinese sentences to obtain parallel data.
Due to the tendency of machine translation to use simple vocabulary and sentence structures, there may be a decrease in text difficulty during both the translation and English simplification stages.

To ensure data quality, we filter the obtained parallel data from three aspects: simplicity, fluency, and semantic retention.
For simplicity, we calculate the average lexical difficulty level for both the original sentence and the simplified sentence using HSK.
Only when the difficulty level of the simplified sentence is significantly reduced compared to the original sentence, this parallel sentence pair will be retained. 
For fluency, we calculate the perplexity for the simplified sentences and filter out sentences above the preset threshold.
For semantic retention, we use sentence-transformers toolkit \cite{reimers2020making} to calculate the semantic similarity between the original sentence and simplified sentence, and also filter out sentences that exceed the preset threshold.

By filtering more than 4 million sentence pairs, we finally get 691,474 sentence pairs that could be used for training. 
By randomly selecting and manually observing 50 entries from these data, we find that the generated sentences had significant simplification operations compared to the original sentences.
Using the filtered data, we train a Chinese BART-base model.

\paragraph{Large Language Models}
We choose two advanced large language models to conduct experiments: \textit{gpt-3.5-turbo} and \textit{text-davinci-003}. 
Both of them are based on GPT-3.5.
The former is the most capable GPT-3.5 model and is optimized for chatting.
The latter is the previous model, which can execute any language task according to instructions.
We translate the simplification prompt used by Feng et al. \shortcite{feng:23} as our prompt.
More details about the prompt can be found in Table \ref{table2}.
Recent studies have shown that the in-context few-shot method can enhance the performance of LLM \cite{brown2020language, wei2022chain}. To further investigate the ability of LLM on Chinese text simplification tasks, we tested the 3-shot performance of the \textit{gpt-3.5-turbo} method by adding 3 instances at the end of our prompt, which is named \textit{gpt-3.5-turbo-few-shot}.

\begin{table}[h]
\begin{center}
\small
\begin{tabular}{|l|}
\hline 
    Our Prompt  \\
    \hline
    我想让你把我的复杂句子替换成简         \\ 
    单的句子。你要保持句意不变，但         \\
    使句子更简单。                      \\
    复杂句： \{Complex Sentence\}        \\
    简单句： \{Simplified Sentence(s)\} \\
\hline
\end{tabular}
\end{center}
\caption{\label{table2} Prompt for Chinese text simplification }
\end{table}

\subsection{Automatic Metrics} \label{5.2}
Following previous work, we choose three metrics for evaluation: SARI \cite{xu:16}, BLEU \cite{papineni:02} and HSK Level \cite{kong2022multitasking}.

\paragraph{SARI}
SARI \cite{xu:16} is a commonly used evaluation metric for text simplification.
Comparing system outputs to multiple simplification references and the original sentences, SARI calculates the mean of the n-gram F1 scores of \textit{add}, \textit{keep}, and \textit{delete}.
In our experiment, we tokenize sentences using Stanford CoreNLP\footnote{\url{https://github.com/stanfordnlp/CoreNLP}} and use the EASSE toolkit \footnote{\url{https://github.com/feralvam/easse}} \cite{alva-manchego-etal-2019-easse} to calculate SARI.

\paragraph{BLEU}
BLEU (Bilingual Evaluation Understudy) \cite{papineni:02} was initially used to evaluate the quality of machine translation.
By calculating the N-gram and counting the times that can be matched, 
BLEU can reflect the closeness between system outputs and references.
Just like calculating SARI, we use the EASSE toolkit to calculate the BLEU score.
    
\paragraph{HSK Level}
In order to measure the complexity of Chinese sentences, we import HSK Level.
HSK is the Chinese proficiency test designed for non-native speakers.
It provides a vocabulary of nine levels from easy to difficult, which includes over 6,000 entries, covering the most common Chinese words.
In our proposed dataset (source sentences), $85.14\%$ of the words could be found in the HSK list, with many of the rest being proper nouns. 
Following previous work \cite{kong2022multitasking}, we count the proportion of words at levels 1-3 and
7+ in system outputs. 
The higher the proportion of words in levels 1-3 (7+), the easier (more challenging) the outputs are understood.
Our specific implementation of this metric is the same as Kong et al. \shortcite{kong2022multitasking}.

\begin{table*}[t]
   \setlength{\tabcolsep}{10pt }
    \renewcommand\arraystretch{1.25}
 
\begin{center}
\begin{tabular}{c|cccc}
\hline
    Method & SARI $\uparrow$ & BLEU $\uparrow$ & L1-3 (\%) $\uparrow$ & L7+ (\%) $\downarrow$ \\
    \hline
    Source & 22.37 & 84.75 & 40.24 & 44.90 \\
    Gold Reference & 48.11 & 61.62 & 46.25 & 39.50 \\
    \hline
    Direct Back Translation & 40.37 & 48.72 & 39.19 & 45.44 \\
    Translated Wiki-Large & 28.30 & \textbf{82.20} & 40.32 & 44.92 \\
    Cross-Lingual Pseudo Data & 38.49 & \underline{63.06} & 41.57 & 44.24 \\
    \hline
    gpt-3.5-turbo-few-shot & \textbf{43.95} & 56.46 & \textbf{44.44} & \textbf{40.67} \\
    gpt-3.5-turbo & \underline{42.39} & 49.22 & \underline{43.68} & \underline{41.29} \\
    text-davinci-003 & 37.97 & 36.18 & 38.80 & 45.32 \\
    
\hline
\end{tabular}
\end{center}
\caption{\label{table3}  The automatic evaluation results on the test set of MCTS. Source refers to the score calculated by taking the original sentence directly as the output of the system. $\uparrow$ The higher, the better. $\downarrow$ The lower, the better. \textbf{Bold} means the best result, and \underline{underline} means the second-best result.}
\end{table*}

\subsection{Human Evaluation}
In order to obtain more comprehensive evaluation results, we further conduct human evaluation.
Following the previous work \cite{dong2019editnts,kumar:20}, we evaluate the Chinese text simplification systems on three dimensions:

\begin{itemize}
    \item Fluency: Is the output grammatical?
    \item Adequacy: How much meaning from the original sentence is preserved?
    \item Simplicity: Is the output simpler than the original sentence?
\end{itemize}

We provide simplifications generated by different systems for the recruited volunteers.
We ask the volunteers to fill out a five-point Likert scale (1 is the worst, 5 is the best) about these simplifications for each dimension. 
Additionally, following Feng et al.'s work  \shortcite{feng:23}, we measure the volunteers' subjective choices by ranking the simplifications to focus on actual usage rather than evaluation criteria.

\section{Results}
In this section, we report the evaluation results of the test set of MCTS. 

\begin{table*}[t]
\begin{center}
    \renewcommand\arraystretch{1.25}
\begin{tabular}{c|ccccc}
\hline
    Method & Simplicity $\uparrow$ & Fluency $\uparrow$ & Adequacy $\uparrow$ & Avg. $\uparrow$ & Rank $\downarrow$ \\
    \hline 
    Direct Back Translation   & 3.42 \footnotesize$\pm$0.87 & 4.36 \footnotesize$\pm$0.78 & \textbf{4.72} \footnotesize$\pm$0.56 & 4.17 & 2.88 \\
    Cross-Lingual Pseudo Data & 4.11 \footnotesize$\pm$0.81 & \underline{4.46} \footnotesize$\pm$0.65 & 3.88 \footnotesize$\pm$0.96 & 4.15 & 2.86 \\
    gpt-3.5-turbo             & \underline{4.17} \footnotesize$\pm$0.89 & \underline{4.46} \footnotesize$\pm$0.70 & \underline{4.43} \footnotesize$\pm$0.78 & \underline{4.35} & \underline{2.29} \\\hline
     Gold Reference            & \textbf{4.20} \footnotesize$\pm$1.08 & \textbf{4.68} \footnotesize$\pm$0.55 & 4.31 \footnotesize$\pm$0.93 & \textbf{4.40} & \textbf{1.97} \\ 
    
\hline
\end{tabular}
\end{center}
\caption{\label{table4}  The human evaluation results. Avg. means the average score of fluency, adequacy, and simplicity. Rank means the subjective ranking for the simplifications. Each annotator annotates all 30 sentences. The values in the table after the $\pm$ symbol are the standard deviation.} 
\end{table*}

\subsection{Results of Automatic Evaluations}
The results of automatic evaluations are shown in Table \ref{table3}.
In addition to the model results, we also report the score of the source and gold reference.
The source scores are calculated on the unedited original sentence.
We calculate the gold reference scores by evaluating each reference against all others in a leave-one-out scenario and then averaging the scores.

To our surprise, direct back translation gets the best SARI score among the unsupervised methods.
But regarding HSK level, the performance of direct back translation is not good, even worse than the source.
We find that many rewrite operations were generated during the back translation process, which is highly correlated with the SARI score.
However, due to the lack of control over simplicity, direct back translation is more like a sentence paraphrase method than text simplification. 
This may be why it performs poorly on the HSK level.

The translated Wiki-Large method gets the best BLEU score but the lowest SARI score among all methods.
In fact, the system output has hardly changed compared to the original sentence.
As the unedited source gets the highest BLEU score of 84.75, we believe the single BLEU value cannot be used as an excellent indicator of text simplification.
Because there is a significant overlap between the original sentence and the references.
As for the poor performance of the translated Wiki-Large method, we believe it is due to the large amount of noise contained in the translated training data.

The SARI score of the cross-lingual pseudo data method is 38.49, which is between the other two unsupervised methods. 
But it performs better on the HSK level than the other two.
This may be because the model learned simplification knowledge from pseudo data that was transferred from the English text simplification model.

In terms of the large language models, the gpt-3.5-turbo significantly performs better than text-davinci-003.
However, although the gpt-3.5-turbo-few-shot under the few-shot setting can further improve performance, it is still insufficient compared to the gold reference.

\subsection{Results of Human Evaluations}

We conduct human evaluations on three representative methods, namely direct back translation, cross-lingual pseudo data, and gpt-3.5-turbo.
We recruit three volunteers to conduct the evaluation.
All of them have a background in linguistics.
We select 30 sentences from the test set of MCTS for each volunteer and provide them with the original sentences and the outputs of these methods.
For the convenience of comparison, a randomly selected reference for each sentence is additionally provided.
Volunteers are asked to rate the simplification of these four groups.
The results of the human evaluation are shown in Table \ref{table4}.

We can see that the gold reference gets the best average score and rank.
It is significantly superior to the output results of other simplification systems.
For detail, it gets the best simplicity score of 4.20 and the best fluency score of 4.68.
Due to some degree of sentence compression, it does not achieve the best adequacy score but only 4.31.
As for the direct back translation method, despite its excellent performance in adequacy, it achieves the lowest simplicity score due to the lack of corresponding control measures. 
On the contrary, the cross-lingual pseudo-data method performs well in terms of simplicity but does not perform well in terms of adequacy.
Because it tends to perform more sentence compression, which removes lots of semantic information.
These two unsupervised methods get a similar average score and rank.

The gpt-3.5-turbo gets the second-best results among all metrics. 
By analyzing the average score and the rank score, we can find that it is significantly better than the two unsupervised simplification methods.
But compared to the gold reference,  there is still a certain gap.
Our experiment has shown that there is still room for further improvement in the large language model's Chinese text simplification ability.

By comparing the results of automatic evaluation and manual evaluation, we can find that the SARI and HSK Level results fit well with the manual assessment rankings.
Methods that performed better on these two metrics also received higher rankings in the human evaluation, while the BLEU scores were almost independent of simplification quality.
Thus we consider SARI and the HSK Level to be valid in assessing the quality of Chinese text simplification.

Besides, several features can be observed in the statistical results of Simplicity, Fluency, and Adequacy:

(1) Gold reference has the highest standard deviation in the simplicity statistics. This suggests that manual simplification may not always tend to be as simple as possible, and fluctuates more in simplicity compared to automatic simplification methods.

(2) On fluency, gold reference receives the highest score and lowest standard deviation, suggesting that gold reference not only strives to keep sentences fluent but also maintains a high level of consistency.

(3) Gold reference and cross-lingual pseudo data obtained large standard deviations and low scores in adequacy. This suggests that these two methods may tend to perform more deletion operations.

\section{Conclusion}
In this paper, we introduce the MCTS, a human-annotated dataset for the validation and evaluation of Chinese text simplification systems. 
It is a multi-reference dataset that contains multiple rewriting transformations.
By calculating the low-level features for simplifications, we have shown the rich simplifications in MCTS, which may be of great significance for understanding the simplification and readability of Chinese text from a linguistic perspective.
Furthermore, we test the Chinese text simplification ability of some unsupervised methods and advanced large language models using the proposed dataset.
We find that even advanced large language models are still inferior to human simplification under the zero-shot and few-shot settings.
Finally, we hope our work can motivate the development of Chinese text simplification systems and provide references for future research.

\section{Acknowledgments}
This work was supported by the MOE (Ministry of Education in China) Project of Humanities and Social Sciences (No. 23YJCZH264), the funds of Research Project of the National Language Commission (No. ZDA145-17) and  Research Funds of Beijing Language and Culture University (No. 23YCX131).







\nocite{*}
\section{Bibliographical References}\label{sec:reference}

\bibliographystyle{lrec-coling2024-natbib}
\bibliography{lrec-coling2024}

\section{Language Resource References}
\label{lr:ref}
\bibliographystylelanguageresource{lrec-coling2024-natbib}
\bibliographylanguageresource{languageresource}

\end{CJK*}
\end{document}